\documentclass[conference]{IEEEtran}
\IEEEoverridecommandlockouts
\usepackage{cite}
\usepackage{amsmath,amssymb,amsfonts}
\usepackage{graphicx}
\usepackage{textcomp}
\usepackage{xcolor}
\usepackage[ruled,linesnumbered,commentsnumbered]{algorithm2e}  
\usepackage{amsmath}
\usepackage{mathrsfs}
\usepackage{bbding}
\usepackage{booktabs}
\usepackage{multirow}
\usepackage[colorlinks, linkcolor=blue, anchorcolor=blue, citecolor=blue]{hyperref}
\usepackage[misc]{ifsym}
\usepackage[normalem]{ulem}
\useunder{\uline}{\ul}{}
\usepackage{subfigure}  

\def\BibTeX{{\rm B\kern-.05em{\sc i\kern-.025em b}\kern-.08em
    T\kern-.1667em\lower.7ex\hbox{E}\kern-.125emX}}
\begin{document}

\title{Source-free Active Domain Adaptation for Diabetic Retinopathy Grading Based on Ultra-wide-field Fundus Image\\
\thanks{This work was supported by the Research Funds of Shanxi Transformation and Comprehensive Reform Demonstration Zone(Grant No. 2018KJCX04), the Fund for Shanxi "1331 Project" and Supported by Fundamental Research Program of Shanxi Province(No.202203021211006. The work was also partially sponsored by the the Key Research and Development Program of Shanxi Province(No.201903D311009).}}

\author{
    \IEEEauthorblockN{Jinye Ran$^{a}$, Guanghua Zhang$^{b,d,*}$\thanks{$^{*}$ is corresponding author.}, Ximei Zhang$^c$, Juan Xie$^d$, Fan Xia$^e$, Hao Zhang$^{f,*}$}
    \IEEEauthorblockA{$^a$ College of Computer and Information Science, Southwest University Chongqing, China}
    \IEEEauthorblockA{$^b$ School of Big Data Intelligent Diagnosis and Treatment industry, Taiyuan University, Taiyuan, China}
    \IEEEauthorblockA{$^c$ College of Biomedical Engineering, Taiyuan University of Technology, Taiyuan, China}
    \IEEEauthorblockA{$^d$ Shanxi Eye Hospital, Taiyuan, China}
    \IEEEauthorblockA{$^e$ Reading Academy, Nanjing University of Information Science and Technology, Nanjing, China}
    \IEEEauthorblockA{$^f$ College of Chemisitry and Chemical Engineering, Southwest University Chongqing, China}
    \IEEEauthorblockA{mac001@126.com, haozhang@swu.edu.cn}
}

\maketitle

\begin{abstract}
Domain adaptation (DA) has been widely applied in the diabetic retinopathy (DR) grading of unannotated ultra-wide-field (UWF) fundus images, which can transfer annotated knowledge from labeled color fundus images. However, suffering from huge domain gaps and complex real-world scenarios, the DR grading performance of most mainstream DA is far from that of clinical diagnosis. To tackle this, we propose a novel source-free active domain adaptation (SFADA) in this paper. Specifically, we focus on DR grading problem itself and propose to generate features of color fundus images with continuously evolving relationships of DRs, actively select a few valuable UWF fundus images for labeling with local representation matching, and adapt model on UWF fundus images with DR lesion prototypes. Notably, the SFADA also takes data privacy and computational efficiency into consideration. Extensive experimental results demonstrate that our proposed SFADA achieves state-of-the-art DR grading performance, increasing accuracy by 20.9\% and quadratic weighted kappa by 18.63\% compared with baseline and reaching 85.36\% and 92.38\% respectively. These investigations show that the potential of our approach for real clinical practice is promising.
\end{abstract}


\section{Introduction}
Diabetic retinopathy (DR) is one of the most commonly occurring secondary microvascular complications in diabetes mellitus, resulting in vision impairment or blindness in at least one eye \cite{wang2018diabetic}. As DR is nearly asymptomatic when it causes irreversible damage to the blood vessels at the back of the eye, early detection and treatment are of utmost importance for preserving the vision of diabetic patients \cite{fong2004retinopathy,gupta2022diabetic,ellis2013teaching}. The most effective way to diagnose DR is to perform regular fundus condition assessment in diabetic patients, which relies on the empirical observation of microaneurysm (MA), hemorrhage (HE), soft exudate (SE), and hard exudate (EX) in fundus images by professional ophthalmologists. To develop an optimal treatment plan, DR can be differentiated into two major classes: non-proliferative DR and proliferative DR, and each fundus image can be specifically divided into five DR stages of progressive severity: normal, mild, moderate, severe non-proliferative, and proliferative according to international protocols \cite{atwany2022deep,ansari2022diabetic}. However, there are precise and repeatable diagnostic criteria for DR grading, but the identification of retinopathy areas and categories is so difficult that it faces problems of inefficiency, subjectivity\cite{srinivasan2023inter}, and even the risk of misdiagnosis.

In recent years, computer-aided diagnosis (CAD) technology based on neural networks has gradually been applied to the task of DR grading based on traditional color fundus images, which can greatly alleviate the anxiety of ophthalmologists. It is reported that these advanced CAD systems achieve DR grading performance equal to or even exceeding that of humans on some real-world DR datasets, such as Eyepacs \cite{graham2015kaggle}. Currently, with the successful commercialization of ultra-wide-field (UWF) fundus photography, DR grading is entering the time of UWF fundus images. Compared with traditional color fundus images covering 30$^{\circ}$-60$^{\circ}$ of retina, the UWF fundus images can capture fundus information from a larger field of view ranging 80$^{\circ}$ to 200$^{\circ}$, which is conducive for early detection of DR \cite{ju2020bridge}. However, due to the difficulty of both acquiring and labeling UWF fundus images, there are currently no available UWF fundus datasets comparable in scale to traditional color fundus datasets. As a result, most CAD systems based on supervised learning paradigms hardly benefit from the advances of photography technology.

To remedy this defect and obtain better performance in DR grading, a large number of DR grading studies based on UWF fundus images have been reported in the research community. Some studies are to annotate the UWF fundus images and to propose a new CAD system, following the previous research paradigm of supervised learning, such as \cite{falavarjani2016ultra, oh2021early}. However, the cost of annotation is so expensive that most attempts are based on a few hundred data points. While better performance in DR grading can be obtained, there is always a risk of overfitting owing to the scarcity of samples for neural network training. Some studies are to switch the research paradigm to transfer learning, such as domain adaptation (DA) \cite{gupta2023learning, wei2023cross}, which enables automatic DR grading of UWF fundus images by transferring the annotated knowledge from the traditional color fundus images domain to the almost unannotated UWF fundus images domain. However, due to the existence of huge domain gaps and the quality of fundus images, most DA methods are hard to achieve an ideal DR grading performance. Meanwhile, the absence of more realistic scenario considerations, such as isolated data islands, has also hindered the further development of DA in DR grading. Nonetheless, considering the possibility of continued development in fundus photography techniques, the latter is of more research value than the former. So, it is imperative to study a DA method that takes into account both application scenarios and grading performance.

In this paper, we propose a novel source-free active domain adaptation (SFADA) for DR grading based on UWF fundus images, focusing on data privacy, computational efficiency, and superior DR grading performance. We aim to extract the DR grading knowledge from a well-trained model of traditional color fundus images and then employ this knowledge to obtain a robust and powerful DR grading model of UWF fundus images. Notably, the proposed framework yields better DR grading performance in more realistic application scenarios, which further improves the practical application prospect of DA in automatic DR diagnosis. The main contributions of this paper are as follows:

\begin{itemize}
\setlength{\itemsep}{0pt}
\setlength{\parsep}{0pt}
\setlength{\parskip}{0pt}
\item[$\bullet$] We propose to model the continuously evolving relationships between different DR grades of color fundus images for fine-grained problems and generate the features of color fundus images by noise for source-free DA, which can take data privacy and computational efficiency into account.
\item[$\bullet$] We propose a novel active local representation matching for valuable UWF fundus images selecting. This method can effectively reduce the impact of outliers on sample actively selecting process and is suitable for the source-free DA scenario of real-world fundus images.
\item[$\bullet$] We propose a novel lesion-based prototype DA to fully utilize UWF fundus images. Models with better DR grading performance are adapted by domain contrastive alignment, domain consistency regularity, and selecting pseudo label with mixup samples.
\item[$\bullet$] Extensive experimental results show that the proposed SFADA achieves state-of-the-art performance in DR grading under the premise of ensuring data privacy and computational efficiency.
\end{itemize}

The rest of the paper is organized as follows: The related work of this paper is reviewed in Section \ref{Related work}. The details of our proposed SFADA framework are elaborated in Section \ref{Methods}. The experimental particulars and results are presented in Section \ref{Experiments}, along with the discussion and the ablation experiments for each part of the method, and we conclude this paper in Section \ref{Conclusion}.

\section{RELATED WORK} \label{Related work}
Our work is closely related to domain adaptation, source-free domain adaptation, and active learning for DR grading. A brief introduction to these three aspects will be presented in this section.

\begin{figure*}[!t]
\centering
\includegraphics[scale=0.53]{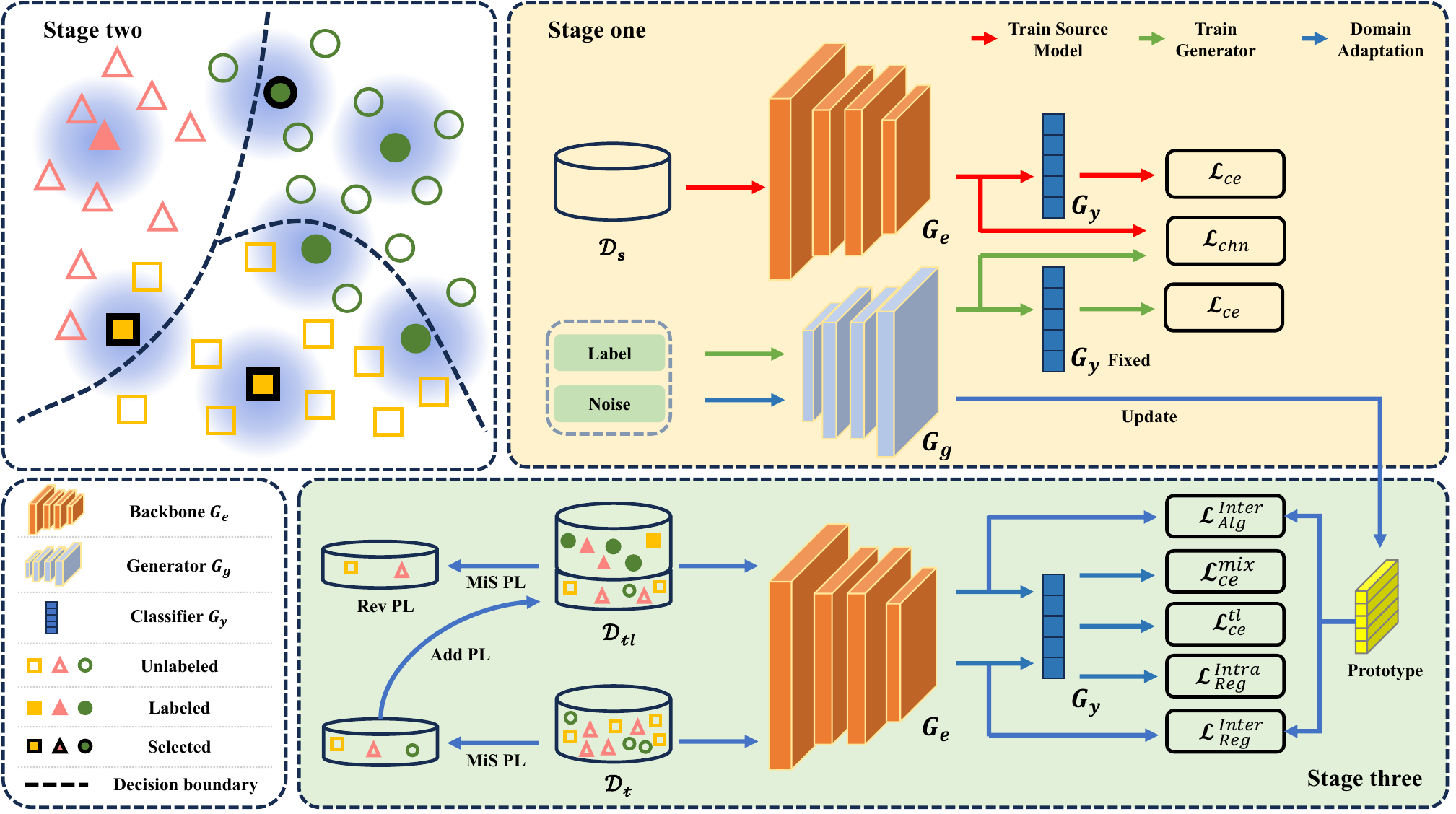}
\caption{The flowchart of the proposed source-free active domain adaptation (SFADA). The Stage one is to train a source pre-trained model with annotated color fundus images and a source feature generator by noise. The Stage two aims to actively select a few valuable ultra-wide-filed (UWF) fundus images for labeling, and the Stage three adapts the source pre-trained model to the UWF fundus images. Notably, The Stage two and the Stage three run alternately.}
\label{flowchart}
\end{figure*}

\subsection{Domain adaptation for DR grading}
Domain adaptation has been widely studied in the field of medical images, which can significantly reduce the dependence of research on annotated fundus images. There are also a lot of related studies based on domain adaptation methods in terms of automatic DR diagnosis. \cite{ju2020bridge} proposed an adversarial domain adaptation to bridge the domain between UWF fundus images and traditional color fundus images. With the help of pseudo-labeling and adversarial learning, the performance of DR grading is improved. \cite{bai2022unsupervised} focused on interpretability in DR grading and investigated an unsupervised lesion-aware transfer learning framework for DR grading in UWF fundus images. A lesion external attention module is introduced to transfer ﬁne lesion knowledge, obtaining a significant improvement in DR grading performance. Inspired by the success of fixed-ratio based mixup in unsupervised domain adaptation, \cite{wei2023cross} designed cross-domain collaborative learning for recognizing multiple retinal diseases. By employing Transformers for producing scale-invariant features and limited annotation of the target domain, more meaningful knowledge transfer is achieved. However, although the methods mentioned above have improved the performance of DR grading compared with their baseline, from the perspective of DR diagnosis, the DR grading performance needs to be further improved.

\subsection{Source-free domain adaptation for DR grading}
Source-free domain adaptation is a new setting of domain adaptation proposed in recent years, considering the data privacy issues in real scenarios. In the field of DR grading, data privacy is even more involved. \cite{zhang2022diabetic} proposed a source-free transfer learning approach for DR grading. The source domain samples with similar target domain samples are generated by a generative adversarial network to realize knowledge transfer with no source domain data. Different from the above methods, \cite{zhou2022domain} argued to generate source domain samples directly by a generative adversarial network and then begin to adapt based on these forged source domain data with label and target domain data. \cite{pourreza2023open} study the source-free domain adaptation in open-set. The source classifier is fixed, while fine-tuning is performed on the source backbone using target domain data to achieve source-free domain adaptation. However, these methods are hard to further develop due to problems such as high computational costs and difficulty accounting for source domain knowledge.

\subsection{Active learning for DR grading}
Active learning is a paradigm that focuses on human-computer interaction and is often used to model complex data in real-world scenarios. Active learning provides a means to leverage expert knowledge to intervene in the neural network modeling process, generally leading to better performance in automatic DR grading. \cite{qureshi2021diabetic} proposed an active learning method named expected gradient length to realize DR grading. With the help of clinical specialists, it achieved remarkable DR grading results on the Eyepacs. \cite{ahsan2022active} proposed a hybrid model for active learning. By quantifying the uncertainty and selecting the worthiest sample to label, nearly perfect area under the curve performance was obtained. However, active learning is still rare in the DR grading field of UWF fundus images.

Inspired by the work mentioned above, this paper proposes a novel SFADA for DR grading of UWF fundus images, focusing on data privacy, computational efficiency, and superior DR grading performance.

\section{PROPOSED METHOD} \label{Methods}
\subsection{Problem formulation}
In the SFADA setting, there is a source pre-trained model $G= \left\{G_e, C_y\right\}$ trained by annotated color fundus images and a target domain consisting of unannotated UWF fundus images $\mathcal{D}_t=\left\{x_i^t\right\}$. Source domain consisting of annotated color fundus images is unavailable during domain adaptation. Meanwhile, a certain amount of unannotated UWF fundus images $\mathcal{B}$ will be actively selected for labeling ($\mathcal{B} \ll \mathcal{D}_t$). Let all of the selected target domain data be $\mathcal{D}_{tl}=\left\{x_i^{tl}, y_i^{tl}\right\}$. The primary goal of SFADA is to adapt $G$ to the target domain with $\mathcal{D}_t$ and $\mathcal{D}_{tl}$, thereby improving the DR grading performance of $G$ in UWF fundus images. 

\subsection{Overall scheme}
We exploit to study a superior performance domain adaptation for UWF-based DR grading, which is closer to the real-world scenarios. As shown in Fig. \ref{flowchart}, the proposed SFADA consist of three stages: source domain feature generation, active local representation matching, and lesion-based prototype domain adaptation. After stage one, stage two and stage three will run alternately for several rounds until the algorithm ends in accordance with the established hyper-parameter settings. More specific details refer to Algorithm \ref{alg::DA}. 

\begin{algorithm}[ht]
  \caption{Overall Scheme of SFADA} % 名称
  \label{alg::DA}
  \begin{footnotesize}  
    \SetAlgoLined
    \SetKwInOut{Input}{Input}\SetKwInOut{Output}{Output}
    \Input{Source pre-trained model $\left\{G_e, C_y\right\}$; Target domain data $\mathcal{D}_t$; Source domain data $\mathcal{D}_s$;  Generator $G_g$; Training epochs $K$,$E$,$M$; Number classes of DR grades $K$;}
    \Output{Adapted model $\left\{G_e, C_y\right\}$}
    \BlankLine
    Initialization $\mathcal{D}_{tl}=\emptyset$; $\mathcal{D}_{tu}=\mathcal{D}_t$; Source prototype $P_k$\;
    \For{$k \leftarrow 0$ \KwTo $K$}{
    Randomly sample a mini-batch from $\mathcal{D}_s$\;
    Update parameters of $\left\{G_e, C_y\right\}$ with Eq. (\ref{ce_loss}) and (\ref{gen_chain_loss})\;
    }
    Freeze parameters of $G_y$\;
    \For{$e \leftarrow 0$ \KwTo $E$}{
    Randomly sample $\xi \sim \mathcal{N}(0,1)$, $\kappa \sim U(0, K)$\;
    Update parameters of $G_g$ with Eq. (\ref{ce_loss}) and (\ref{gen_chain_loss})\;
    }
    Freeze parameters of $G_g$\;
    \For{$m \leftarrow 0$ \KwTo $M$}{
    \If{need active selection}{
    Obtain active selection set $\mathcal{D}_a$ as described in Algorithm \ref{alg::AL}\;
    $\mathcal{D}_{tl} \leftarrow \mathcal{D}_{tl} \cup \mathcal{D}_a$, $\mathcal{D}_{tu} \leftarrow \mathcal{D}_{tu} \setminus \mathcal{D}_a$\;}
    Initialize $\mathcal{D}_{add}=\emptyset$, $\mathcal{D}_{rev}=\emptyset$, $\mathcal{S} \leftarrow \mathcal{D}_{tl}$, iterations $I$ on $\mathcal{D}_{tu}$\;
    \For{$i \leftarrow 0$ \KwTo $I$}{
    Update source prototype $P_k$ with $G_g$ and Eq. (\ref{ema})\;
    Sample mini-batches $\left\{x_{tl},y_{tl}\right\}$ from $\mathcal{S}$, $x_{t}$ from $\mathcal{D}_{tu}$\;
    Calculate $\mathcal{L}_{ce}^{tl}$, $\mathcal{L}_{Alg}^{Inter}$, $\mathcal{L}_{Reg}^{Inter}$, $\mathcal{L}_{Reg}^{Intra}$, $\mathcal{L}_{ce}^{mix}$ with Eq. (\ref{tgtl_ce}), (\ref{alg_inter}), (\ref{inter_reg}), (\ref{intra_reg}), (\ref{ce_mix}), and Update the parameters of $\left\{G_e, C_y\right\}$\;
    \If{need pseudo label}{
    Update $\mathcal{D}_{add}$ and $\mathcal{D}_{rev}$ as described in Algorithm \ref{alg::RE_PL}\;
    \If{iteration over $\mathcal{S}$ finishes}{
    $\mathcal{S} \leftarrow \mathcal{S} \setminus \mathcal{D}_{rev} \cup \mathcal{D}_{add}$, $\mathcal{D}_{add}=\emptyset$, $\mathcal{D}_{rev}=\emptyset$\;}}}}
  \end{footnotesize}
\end{algorithm}

\subsection{Source domain feature generation}
Source-free domain adaptation is a successful attempt to respond to the data privacy of DR grading \cite{zhou2022domain, zhang2022diabetic}. However, most mainstream methods suffer from the computational efficiency, which is not conducive to further development. To alleviate this, inspired by \cite{qiu2021source}, we propose a novel source domain feature generation for domain adaptation without source data. Specifically, our approach is divided into two steps. The first step is to obtain a source pre-trained model $\left\{G_e, G_y\right\}$ with color fundus images via a cross-entropy loss function

\begin{footnotesize}  
\begin{equation}
\mathcal{L}_{ce}=\mathbb{E}_{(x,y)\sim\mathcal{D}_s}-\sum_{k=0}^Ky_klog(C_y(G_e(x)))
\label{ce_loss}
\end{equation}
\end{footnotesize}

\noindent where $k$ is one hot label. The second step is to train a source feature generator $G_g$ using a normal noise $\xi \sim \mathcal{N}(0,1)$ and a uniform label $\kappa \sim U(0, K)$ under the condition of a source classifier $G_y$, which loss function is still $\mathcal{L}_{ce}$. Notably, instead of generating source-like samples, our method generates source features for source-free domain adaptation, which can improve computational efficiency.

To further control the computational cost of source-free domain adaptation, we re-think the influence of the fundus image resolution on DA. DR grading using lower-resolution fundus images can significantly improve computational efficiency, while some fundus lesion regions (e.g., HE and EX) become so small that fine-grained feature adaptation is required to be considered. To deal with this issue, we propose to model the continuously evolving relationship from normal to proliferative DR. In terms of mutual information, this implies greater mutual information between neighboring DR grades. However, high-dimensional mutual information is difficult to estimate and calculate \cite{belghazi2018mine}. Meanwhile, directly maximizing the lower bound of mutual information will conflict with the optimization objective of the cross-entropy loss function, resulting in its collapse. Fortunately, we can implicitly optimize the mutual information between different DR grades via an improved contrastive loss function. 

\begin{figure}[htbp]
  \centerline{\includegraphics[scale=0.29]{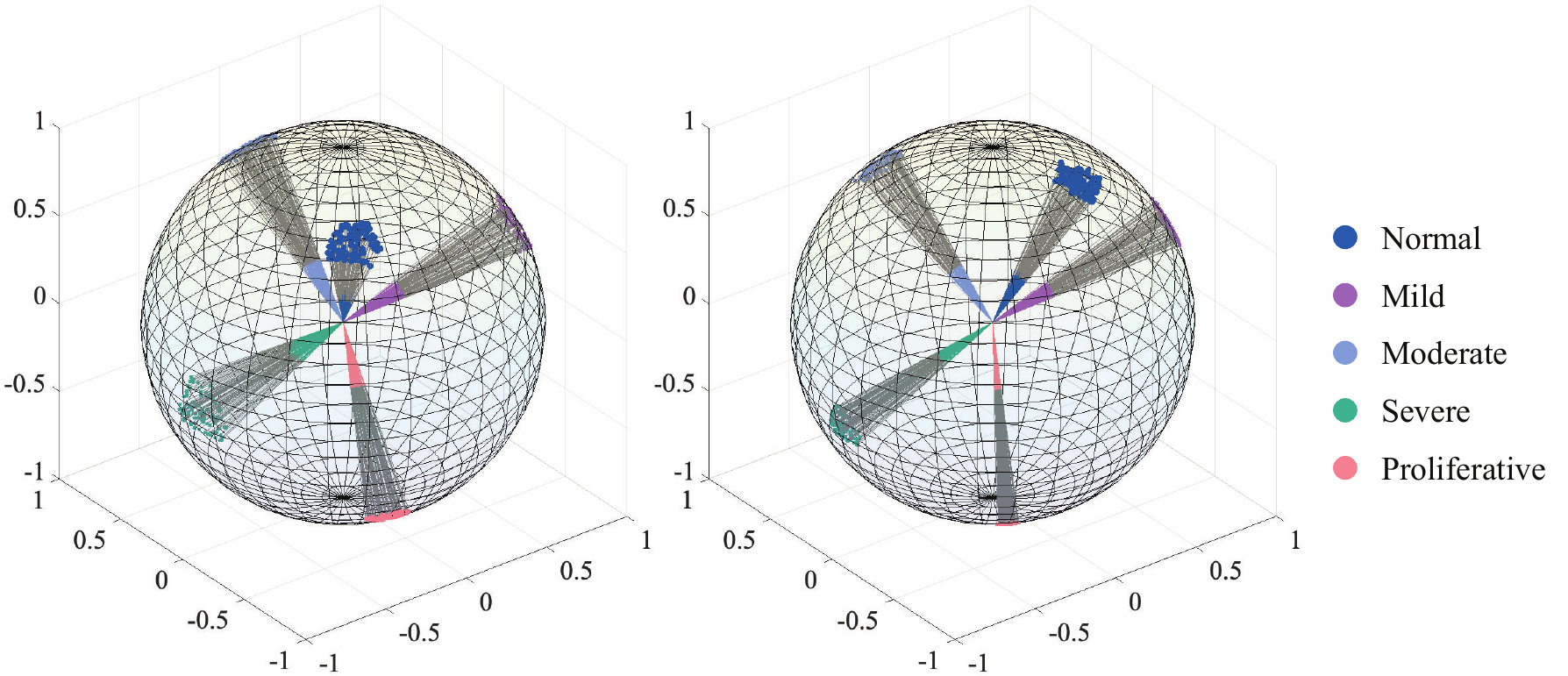}}
  \caption{Visualization of source domain feature in the hyper-sphere. The left is trained by $\mathcal{L}_{ce}$ and the right is trained by $\mathcal{L}_{ce} + \mathcal{L}_{chn}$, comparing the angles and distances between the features of different DRs, which shows that the proposed $\mathcal{L}_{chn}$ successfully models the continuously evolving relationships of DRs.}
  \label{sf}
\end{figure}

\begin{footnotesize}  
\begin{equation}
\mathcal{L}_{chn}=-log\frac{exp((p \cdot k^+-\gamma)/\tau)}{exp((p \cdot k^+ - \gamma)/\tau) + \sum_{j=1}^{C-1}{exp((p \cdot k^- - \beta_j)/\tau)}}
\label{gen_chain_loss}
\end{equation}
\end{footnotesize}

\noindent where $p$, $k^+$, and $k^-$ denote anchor samples, positive samples for anchor sample, and negative samples for anchor sample. $\gamma$, $\tau$, and $\beta$ represent hyper-parameters, respectively. As shown in Fig. \ref{sf}. the effect of the proposed loss function $\mathcal{L}_{chn}$ on color fundus images distribution in the feature space is visualized onto the hyper-sphere. The left hyper-sphere represents the source pre-trained model trained with $\mathcal{L}_{ce}$, and the right one denotes the source pre-trained model trained with $\mathcal{L}_{ce} + \mathcal{L}_{chn}$. Obviously, comparing the angles and distances between features of different class, we can observe that $\mathcal{L}_{chn}$ successfully models the relative relationship between different DR grades.

To this end, both the source pre-trained model and the source feature generator are trained with $\mathcal{L}_{ce} $ and $\mathcal{L}_{chn}$. Essentially, $\mathcal{L}_{chn}$ increases the convergence difficulty of $\mathcal{L}_{ce}$ while modeling the continuously evolving relationship from normal to proliferative DR, obtaining a more powerful source classifier $G_y$, which is conducive to train the source feature generator $G_g$ and adapt the fine-grained knowledge. The flowchart of source domain feature generation is illustrated in stage one of Fig. \ref{flowchart}.

\subsection{Active local representation matching}
Guided by uncertainty and representativeness criteria, active learning selects a few valuable UWF fundus images for labeling and supervised training, thus greatly improving the performance of DR grading. However, due to outliers such as eyelash occlusion and huge domain gaps, the general active learning method is difficult to directly apply to DA in real-world UWF fundus images. To handle this, inspired by the exploration of local information in active domain adaptation \cite{sun2022local, li2022source}, we propose an active local representation matching (ALRM) for active sample selection of UWF-based DR grading in DA. 

Let $p(\cdot)=\sigma(\mathbf{W} \otimes h(\cdot))$ denote different DR probabilities after a softmax layer $\sigma(\cdot)$, where $\mathbf{W}$ and $h(\cdot)$ are the parameters of classifier $G_y$ and feature extracted by source backbone $G_e$, respectively. To alleviate the effect of source-free setting and make full use of source pre-trained model, $\mathbf{W}$ is regarded as class prototypes to map low-dimensional semantic information to high-dimensional feature information and the local representation of each UWF fundus image $x_t$ can be measured as

\begin{footnotesize}  
\begin{equation}
LR(x_t) = \frac{1}{K} \sum_{i=1}^{K} h(x_t^i) \odot p(x_t^i) \otimes \mathbf{W}^\top + h(x_t^i)
\label{lr}
\end{equation}
\end{footnotesize}

\noindent where $x_t^i$ is $i$ nearest neighbors of $x_t$, calculating by cosine similarity. There is an insight that local representation is essentially an averaging of the most similar sample features, reducing the influence of outliers on data-sensitive active sample selection. Meanwhile, the introduction of class probabilistic information also reduces the risk of collapse.

Based on obtained all of local representations of $\mathcal{D}_t$, we try to actively find a subset $\mathcal{D}_{tl} \subset \mathcal{D}_t$ such that $\mathcal{D}_{tl}$ matches $\mathcal{D}_t$ as well as possible. Although this is a NP-hard problem that cannot be solved in polynomial time, it could be approximated by a greedy selection process with relatively correct results. Inspired by \cite{hwang2022combating}, an unlabeled UWF fundus image that minimizes the squared Maximum Mean Discrepancy (MMD) between $\mathcal{D}_t$ and $\mathcal{D}_{tl}$ is selected each time, which can be formulated as

\begin{footnotesize}  
\begin{equation}
\mathop{\arg\min}\limits_{x_{t}} M\!M\!D^2(\mathcal{D}_t, \mathcal{D}_{tl} \leftarrow x_{t})
\end{equation}
\end{footnotesize}

\noindent The process of local representation matching is shown in stage two of Fig. \ref{flowchart}. An interesting finding is that the designed subset matching works only with our proposed local representation. After further analysis of UWF fundus images, we speculate that the intra-class differences of different DR grades are too large, and the limited active samples could not fully match the entire feature space, resulting in the deterioration of active learning performance. Additionally, considering the initial poor performance of $G$ on the UWF fundus images, active learning is performed for $K$ rounds, and $\mathcal{B}/K$ UWF fundus images of the target domain are selected in each round. More details refer to Algorithm \ref{alg::AL}. 

\begin{algorithm}[ht]
\caption{Process of ALRM} % 名称
\label{alg::AL}
\begin{footnotesize}  
  \SetAlgoLined
  \SetKwInOut{Input}{Input}\SetKwInOut{Output}{Output}
  \Input{Per-round budget $\mathcal{B}/K$; Target data $\mathcal{D}_{t}$; Labeled target data $\mathcal{D}_{tl}$}
  \Output{selected set $\mathcal{D}_a$}
  \BlankLine
  Initialization selected set $\mathcal{D}_a = \emptyset$\;
  Obtain local representation $LR_t$ of $\mathcal{D}_{t}$ with Eq. (\ref{lr})\;
  \For{$b\leftarrow 0$ \KwTo $\mathcal{B}/K$}{
  $\mathcal{M} \leftarrow \mathcal{D}_{tl} \cup \mathcal{D}_a$\;
  \For{$x_t \in \mathcal{D}_t$}{
  $\mathcal{M} \leftarrow \mathcal{M} \cup x_t$\;
  Obtain local representation $LR_{\mathcal{M}}$ of $\mathcal{M}$ with Eq. (\ref{lr})\;
  $\mathcal{Z}_t \leftarrow M\!M\!D^2(LR_t, LR_{\mathcal{M}})$\;
  $\mathcal{M} \leftarrow \mathcal{M} \setminus x_t$\;
  }
  $\mathcal{D}_a \leftarrow \mathop{\arg\min}\limits_{x_t} \mathcal{Z}_t$ \;
  }
  % \textbf{Return} $\mathcal{D}_a$
\end{footnotesize}
\end{algorithm}

\subsection{Lesion-based prototype domain adaptation}
After actively selecting, domain adaptation is usually implemented using classical DANN \cite{ganin2016domain} or MME \cite{saito2019semi}, while supervised training is performed on the active UWF fundus images. However, the DR grading performance achieved by these methods is far from our expectations. To re-think the advantages of actively selecting samples, we note that these annotated UWF fundus images can be more fully exploited by semi-supervised methods, like consistency regularity \cite{sohn2020fixmatch}. To this end, we propose a lesion-based prototype domain adaptation (LPDA) to adapt $G$ to UWF fundus images better, which is shown in stage three of Fig. \ref{flowchart}. The proposed LPDA includes three parts: prototype-based domain contrastive alignment, inter and intra domain consistency regularity, and selecting pseudo label with mixup samples. 

\subsubsection{Prototype-based domain contrastive alignment}
Although supervised training of actively selecting UWF fundus images is an efficient way to improve the DR grading performance, the updating direction of the model gradient may not be conducive to the annotated knowledge transfer of color fundus images or even harmful. To avoid this, considering lack of color fundus images, we argue for a prototype contrastive loss function for explicit domain alignment. Specifically, the source features generated by $G_g$ are to update a memory-bank as follows

\begin{footnotesize}  
\begin{equation}
\textbf{h}_i \leftarrow \beta \textbf{h}_{i} + (1 - \beta)\textbf{o}_i
\label{ema}
\end{equation}
\end{footnotesize}

\noindent where $h_i$ and $o_i$ denote the source class prototypes in the memory-bank and class features of the current step, and $\beta=0.99$ is a hyper-parameter to control the ratio of class prototypes updating. The effect of the memory-bank is to ensure the stability of the source class prototype. Therefore, when the cross-entropy loss function 

\begin{footnotesize}  
\begin{equation}
\mathcal{L}_{ce}^{tl}=\mathbb{E}_{(x,y)\sim\mathcal{D}_{tl}}-\sum_{k=0}^Ky_klog(C_y(G_e(x)))
\label{tgtl_ce}
\end{equation}
\end{footnotesize}

\noindent is used for supervised training, the prototype contrastive loss can be used to align the features of the annotated UWF fundus images with the source class prototype in the memory-bank. The prototype contrastive loss is formed as

\begin{footnotesize}  
\begin{equation}
\mathcal{L}_{Alg}^{Inter}=-log\frac{exp(h_{tl} \cdot h^+/\tau)}{exp(h_{tl} \cdot h^+/\tau) + \sum_{n=1}^{K-1}exp(h_{tl} \cdot h^- /\tau)}
\label{alg_inter}
\end{equation}
\end{footnotesize}

\noindent where $h_{tl}$, $h^+$, $h^-$, $K$, and $\tau$ denote the feature of $x_{tl}$ extracting by $G_e$, source class prototype with the same label as $x_{tl}$, source class prototype with the different labels as $x_{tl}$, number classes of DR grades, and hyper-parameter.

\subsubsection{Inter and intra domain consistency regularity}
Inspired by consistency regularity in semi-supervised learning, we can further exploit unannotated UWF fundus images and propose to regularize  the inter-domain consistency and intra-domain consistency. Following the previous literature \cite{sohn2020fixmatch}, weak and strong data augmentations are applied to the unannotated UWF fundus image, obtaining two features $h_{xu}^w$ and $h_{xu}^s$ after extracting by source backbone $G_e$. The inter-domain consistency is to transport $h_{xu}^w$ to all of the source class prototypes and get a soft probability. Then, the soft probability can be used as a supervised signal to regularize the model $G$ with $h_{xu}^s$. This loss function can be formed as 

\begin{footnotesize}  
\begin{equation}
\mathcal{L}_{Reg}^{Inter}= \mathcal{L}_{L1}(\sigma(\gamma<h_{xu}^w, p>), G_y(h_{xu}^s))  
\label{inter_reg}
\end{equation}
\end{footnotesize}

\noindent where $\gamma$, $p$, and $\mathcal{L}_{L1}$ denote optimal transport strategy, source class prototypes, and L1-Loss, respectively. The intra-domain consistency is to obtain the class probabilities of $h_{xu}^w$ and $h_{xu}^s$ after the source classifier $G_y$ and a softmax layer, enforcing both class probabilities to have similar probabilistic output. This loss function can be implemented by 

\begin{footnotesize}  
\begin{equation}
\mathcal{L}_{Reg}^{Intra}= p_{tu}^w \otimes {p_{tu}^s}^\top - \mathrm{tr}(p_{tu}^w \otimes {p_{tu}^s}^\top)
\label{intra_reg}
\end{equation}
\end{footnotesize}

\noindent where $p_{tu}^w$ and $p_{tu}^s$ denote the class probabilities of the same UWF fundus image after weak data augmentation and strong data augmentation. With the help of inter-domain consistency regularity and intra-domain consistency regularity, more details that can be used for DR grading in UWF fundus images are captured by the model $G$.

\subsubsection{Selecting pseudo label with mixup samples}
We notice that the fine-grained problem and source-free setting make DA sensitive to the pseudo-labels. It is hard to find a method to deal with this issue after investigating a lot of literature. Fortunately, the annotated UWF fundus images selected by ALRS prompt us to propose a novel method to select pseudo label with mixup samples (S-PMiS). The proposed S-PMiS allows for better DR grading performance after active sample selection. More details, at the beginning of adaptation, the model $G$ is trained with selected UWF fundus images augmented by mixup

\begin{footnotesize} 
\begin{equation}
  \left\{
    \begin{array}{ll}
      \lambda = B(\alpha, \beta) \\
      x_{mix}^{tl} = \lambda \cdot x_i^{tl} + (1 - \lambda) \cdot x_j^{tl} \\
      y_{mix}^{tl} = \lambda \cdot y_i^{tl} + (1 - \lambda) \cdot y_j^{tl} \\
    \end{array}
\right.
\end{equation}
\end{footnotesize}

\noindent where $B$ refers to a $Beta$ distribution. The mixup augmentation can make the model better able to distinguish whether a mixup sample is generated by two same label UWF fundus images, which can be optimized by a cross-entropy loss function

\begin{footnotesize}  
\begin{equation}
\mathcal{L}_{ce}^{mix} = \mathbb{E}_{(x,y)\sim\mathcal{D}_{tl}}-\sum_{k=0}^Ky_{mix}^{tl}log(C_y(G_e(x_{mix}^{tl})))
\label{ce_mix}
\end{equation}
\end{footnotesize}

\noindent When starting to adapt with a pseudo label, one of the samples $x_r$ with the least similarity to the annotated UWF fundus image $ x_{tl}$ is selected to feed to model $G$, getting its corresponding class probability $p_r$ and pseudo label $y_r$. The similarity between UWF fundus images is measured by a cosine metric based on local representation. All candidates that can be selected are formed as 

\begin{footnotesize}  
\begin{equation}
\left\{x_r^1, x_r^2, \cdots, x_r^k  \right\} = TopK(- <LR(x_{tl}), LR(\omega)>)
\label{candi}
\end{equation}
\end{footnotesize}

\noindent where $k$ and $\omega$ denote a hyper-parameter and a feature memory-bank of all UWF fundus images updated by an exponential moving average (EMA) \cite{tarvainen2017mean}, respectively. Notably, $x_r$ is randomly selected among candidates. If the maximum $p_r$ exceeds the threshold $\phi_a$, $x_r$ is pseudo-labeled and mixed up with an annotated UWF fundus image $x_{tl}$ whose label is the same as the pseudo label of $x_r$ to generate a mixup sample $x_{s}$, following as

\begin{footnotesize}  
\begin{equation}
x_{s} = 0.5 \cdot x_{r}^{pl} + 0.5 \cdot x_{tl}^{lb} \quad \mathrm{s.t.} \; pl = lb\\
\label{cf_sample}
\end{equation}
\end{footnotesize}

\begin{table*}[htbp]
\caption{Quantitative results of SFADA compared with other methods. The best performance is bold fonts and the second performance is underlined. All experiments are conducted with an active budget $\mathcal{B}=5\%$. }
\begin{center}
\footnotesize
\setlength{\tabcolsep}{9pt}
\resizebox{\linewidth}{!}{
\begin{tabular}{@{}ccccccc|ccccc@{}}
\toprule
Method   & Source-free & Normal & Mild  & Moderate & Severe & Proliferative   & ACC   & F1-Score & AUC   & Kappa & Q.W.Kappa   \\ \midrule
DUC      &  \XSolidBrush    & 64.05 & 24.80 & \underline{80.90}    & 21.35  & 71.79 & 58.80 & 52.45    & 81.01 & 45.08 & 72.00 \\
EADA     &  \XSolidBrush    & 69.24 & 26.91 & \textbf{89.86}    & 67.42  & 83.33 & \underline{68.60} & 65.45    & 91.05 & 58.72 & 75.00 \\
LADA     &  \XSolidBrush    & 59.11 & \underline{57.52} & 76.80    & \textbf{84.27}  &80.45 & 68.14 & \underline{68.60}    &\underline{92.36} &\underline{59.02} &\underline{81.25} \\
LADMA    &  \XSolidBrush    & 80.80 & 17.94 & 75.24    & 74.72  & 75.96 & 67.31 & 63.20    & 84.13 & 56.06 & 79.75 \\
SDM-AG   &  \XSolidBrush    & \underline{87.47} & 28.76 & 40.93    & 48.31  & 75.64 & 61.33 & 56.67    & 84.40 & 47.27 & 72.19 \\
ENLP     &  \CheckmarkBold  & 61.39 & 47.23 & 18.91    & 75.28  & 82.05 & 53.00 & 50.47    & 84.83 & 39.66 & 72.74 \\
Baseline &  \CheckmarkBold  & 71.01 & 38.00 & 59.84    & 57.30  & \textbf{91.67} & 64.46 & 60.90    & 90.41 & 53.66 & 73.75 \\
SFADA    &  \CheckmarkBold  & \textbf{93.54} & \textbf{74.14} & 80.50    & \underline{78.65}  & \underline{90.06} & \textbf{85.36} & \textbf{83.71}    & \textbf{95.70} & \textbf{80.54} & \textbf{92.38} \\ \bottomrule
\end{tabular}}
\label{compar}
\end{center}
\end{table*}

\noindent where $pl$ and $lb$ denote the pseudo label of $x_r$ and annotation of $x_{tl}$. Then, $x_{s}$ is fed to model $G$ to obtain class probability $p_{s}$ and pseudo label $y_s$ (MiS PL). If the maximum $p_{s}$ is exceeds the threshold $\phi_b$ and $y_s$ is equal to $y_r$, the pseudo label of $x_r$ is considered to be reliable. At the loop end of the dataloader, $x_r$ with its pseudo label $y_{r}$ will be added to $\mathcal{D}_{tl}$ as an annotated UWF fundus image (Add PL). After the first Add PL, all the UWF fundus images with pseudo labels will be re-checked multiple times by the MiS PL. At this time, if the predicted label of $x_{s}$ by model $G$ is different from its corresponding label of $x_{tl}$, the pseudo label $y_r$ is considered to be unreliable, and $x_r$ will be removed from $\mathcal{D}_{tl}$ before the next loop of the dataloader (Rev PL). More details are shown in Algorithm \ref{alg::RE_PL}.

\SetCommentSty{myCommentStyle}
\newcommand\myCommentStyle[1]{\footnotesize\ttfamily\textcolor{blue}{#1}}
\begin{algorithm}[ht]
  \caption{Process of S-PMiS} % 名称
  \label{alg::RE_PL}
  \begin{footnotesize}  
    \SetAlgoLined
    \SetKwInOut{Input}{Input}\SetKwInOut{Output}{Output}
    \Input{Model $G = \left\{G_e,G_y\right\}$; Memory-bank $\omega$; Annotated target data $\left\{x_{tl},y_{tl}\right\}$; Unannotated target data $\left\{x_{tu}\right\}$; $\mathcal{D}_{rev}$; $\mathcal{D}_{add}$; $\mathcal{D}_{tl}$}
    \Output{$\mathcal{D}_{rev}$; $\mathcal{D}_{add}$}
    \BlankLine
    $p_{tu} = softmax(G_y(G_e(x_{tu})))$, $y_{tu} = \arg \max p_{tu}$\;
    Obtain $LR(x_{tu})$ with Eq. (\ref{lr}) and update $\omega$ with EMA\;
    \If(\tcp*[h]{MiS PL}){$x_{tl} \in \mathcal{D}_{add}$}{
    Generate $x_{s}$ with Eq. (\ref{cf_sample}) and obtain pseudo label $y_{s}$ by G\;
    \If(\tcp*[h]{Rev PL}){$y_{s} \neq y_{tl}$}{
    $\mathcal{D}_{rev} \leftarrow \mathcal{D}_{rev} \cup \left\{x_{tl}, y_{tl}\right\}$\;}
    }
    Obtain all candidates with Eq. (\ref{candi}) and randomly select $x_{r}$\;
    $p_r = softmax(G_y(G_e(X_{r})))$, $y_r = \arg \max p_r$\;
    \If(\tcp*[h]{MiS PL}){$p_r[y_r] \geq \phi_a$}{
    Generate $x_{s}$ with Eq. (\ref{cf_sample})\;
    $p_s = softmax(G_y(G_e(X_{s})))$, $y_s = \arg \max p_s$\;
    \If(\tcp*[h]{Add PL}){$p_{s}[y_{s}]>\phi_b$ and $y_{s}$ = $y_{r}$}{
    $\mathcal{D}_{add} \leftarrow \mathcal{D}_{add} \cup \left\{x_r, y_r\right\}$\;}}
  \end{footnotesize}
\end{algorithm}

\subsection{Overall loss function}
Based on the above work, the proposed LPDA is optimized by 

\begin{footnotesize}  
\begin{equation}
\mathcal{L} = \mathcal{L}_{ce}^{tl} +\alpha \cdot \mathcal{L}_{Alg}^{Inter} + \beta \cdot \mathcal{L}_{Reg}^{Inter} + \gamma \cdot \mathcal{L}_{Reg}^{Intra} + \kappa \cdot \mathcal{L}_{ce}^{mix}
\label{over_all}
\end{equation}
\end{footnotesize}

\noindent where $\mathcal{L}_{ce}^{tl}$ is the supervised cross entropy loss (Eq. \ref{tgtl_ce}), $\mathcal{L}_{Alg}^{Inter}$ is the prototype contrastive loss (Eq. \ref{alg_inter}), $\mathcal{L}_{Reg}^{Inter}$ is the inter-domain consistency regularity loss (Eq. \ref{inter_reg}), $\mathcal{L}_{Reg}^{Intra}$ is the intra-domain consistency regularity loss (Eq. \ref{intra_reg}) and $\mathcal{L}_{ce}^{mix}$ is the supervised cross entropy loss (Eq. \ref{ce_mix}) with mixup, respectively. $\alpha$, $\beta$, $\gamma$, and $\kappa$ denote hyper-parameters for different loss terms.

\section{EXPERIMENTS} \label{Experiments}

\subsection{Datasets and metrics}
The public Eyepacs dataset and a private UWF fundus images dataset are employed as the source domain and target domain of our experiments, respectively. Considering overall efficiency and a realistic setup, 6000 color fundus images are randomly selected from Eyepacs to establish a new subset for the training of a source pre-trained model and other comparative methods while preserving the issue of data imbalance. The private dataset consists of 2816 UWF fundus images and has the same DR grade as Eyepacs (i.e., 906 normal, 553 mild, 720 moderate, 247 severe, and 390 proliferative). Meanwhile, to avoid a potential domain shift between the train dataset and test dataset, we follow previous DA practices \cite{sun2022local, qiu2021source}, retaining all of the UWF fundus image for training and testing rather than k-fold cross-validation. For a comprehensive quantitative evaluation, overall and class-specific accuracy (ACC), F1-score, receiver operating characteristic (ROC), area under the curve (AUC), kappa, and quadratic weighted kappa (Q.W.Kappa) are reported in our experimental results.

\subsection{Implementation details}
The resolution of fundus images is resized to 512 $\times$ 512 for memory efficiency. Empirically, the same data augmentation techniques as used in the VisDA2017 \cite{peng2017visda} are applied to augment all of the fundus images, obtaining a resolution of 224 $\times$ 224 for computational efficiency. The source pre-trained model and the source feature generator are trained for 100 epochs and 1000 epochs, respectively, using the same mini-batch size of 64. The source pre-trained model adapts to UWF fundus images for 15 epochs with a mini-batch size of 32. All of the experiments are optimized by Adadelta \cite{zeiler2012adadelta} with default settings except $lr=0.1$. To obtain better performance, the backbone in the source pre-trained model adopt tiny Swim Transformer \cite{liu2021swin}. Meanwhile, the backbone of comparative methods is also replaced for fair comparison. The network of the source feature generator is referred to \cite{xu2020generative} which is initialized by random normal. Our experiments are run on the PyTorch platform with a single RTX3090 GPU. More details refer to \url{https://github.com/JinyeRAN/source-free_active_domain_adaptation}.

\begin{figure}[htbp]
  \centerline{\includegraphics[scale=0.445]{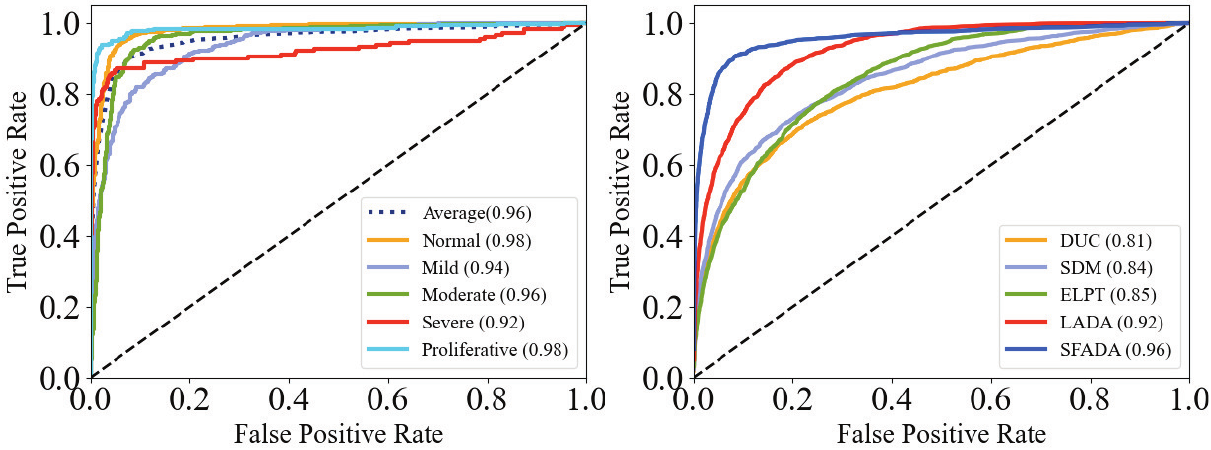}}
  \caption{The receiver operating characteristic curve (ROC) of model trained by SFADA and other methods. The left one is class-specific results of SFADA and the right one is ROC of SFADA and others.}
  \label{roc}
\end{figure}

\begin{figure*} 
\centering    
% \hspace{-3mm}  
\subfigure[Different budget to DR grading.] {
 \label{fig:budget}    
 \includegraphics[width=4.4cm,height=3.34cm]{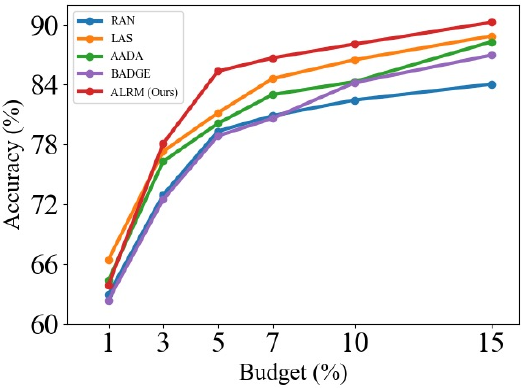} 
}\hspace{-3mm}     
\subfigure[Accuracy of train processing.] { 
\label{fig:budget_fixed}     
\includegraphics[width=4.4cm,height=3.34cm]{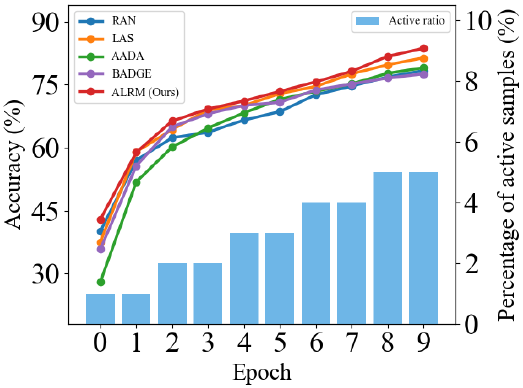}  
}\hspace{-3mm}    
\subfigure[The effect of $K$ to DR grading.] { 
\label{fig:AL_k}     
\includegraphics[width=4.4cm,height=3.34cm]{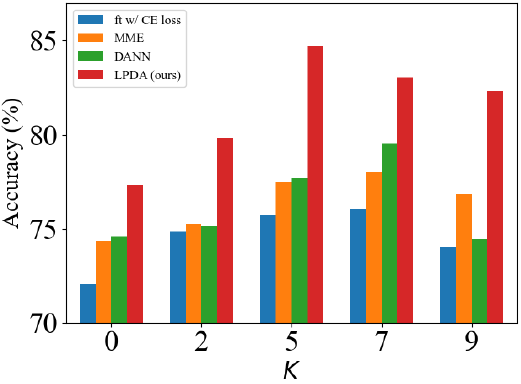}  
}\hspace{-3mm}   
\subfigure[Accuracy of pseudo label] { 
\label{fig:S-RMiS}     
\includegraphics[width=4.4cm,height=3.34cm]{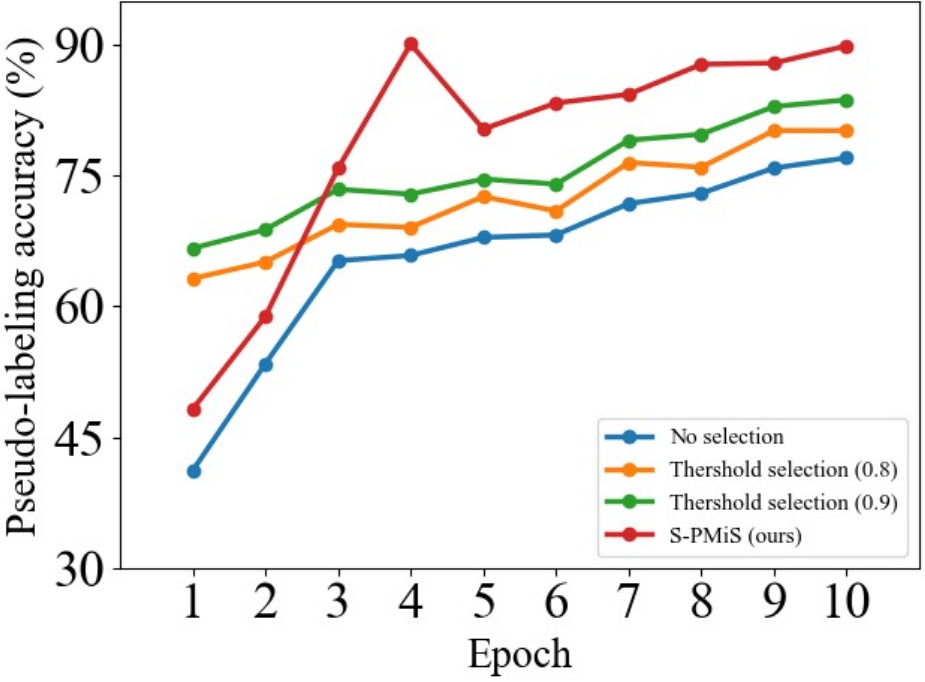}
}
\caption{(a)-(c) Analysis of the impact of active sample selection on the final DR grading performance, where (a) reports the impact of different budgets, (b) reports the accuracy of train processing with a fixed budget $\mathcal{B}$=5\%, and (c) reports the influence of different numbers of nearest neighbors $K$. (d) The influence of the proposed selecting pseudo label with mixup samples on the accuracy of pseudo label.}     
\label{spp}     
\end{figure*}

\subsection{Overall evaluation}
Since our SFADA mainly focuses on implementing a more realistic and superior method in DR grading, it is difficult to find an available baseline for improvement. Therefore, we modify LADA \cite{sun2022local} to make it our baseline to study the effectiveness of our proposed methodology. In addition, due to the lack of a publicly available UWF benchmark dataset, we cannot directly utilize the reported domain adaptation performance of DR grading between color fundus images and UWF fundus images from existing literature. So, all the experimental results in this paper are reproduced by us on our dataset. We choose some methods that are similar to our research background for side-by-side comparison, including DUC \cite{xie2023dirichlet}, EADA \cite{xie2022active}, LADA \cite{sun2022local}, LADMA \cite{hwang2022combating}, SDM-AG \cite{xie2022learning}, and ENLP \cite{li2022source}. All experiments are conducted with an active budget of 5\%. The quantitative evaluation performance is presented in Table \ref{compar}. The highest score is shown in bold font, while the second-highest score is underlined. The comparison results indicate that (1) Our SFADA achieves remarkable performance far beyond other methods in all five overall evaluation metrics, as well as highly competitive results in class-specific accuracy. Specifically, compared to baseline, the proposed method achieves performance improvements of 18.63\% in Q.W.Kappa and 20.90\% in ACC, both of which are crucial for DR diagnosis. (2) Moreover, compared to other DA methods that rely on source data for DR grading, our SFADA can solve the privacy of medical data to some extent. In contrast to other source-free domain adaptation approaches, SFADA obtains better DR grading performance. (3) As for computational efficiency, although it is difficult to give a quantitative evaluation due to fairness, considering that SFADA adopts one of the smallest fundus image resolutions in DR grading, employs the backbone with fewer parameters, and utilizes generating source features to transfer knowledge, we can make a qualitative conclusion that our approach achieves better computational efficiency than most DA-based DR grading methods. To further verify the performance of the adapted model on UWF fundus images, we plot the receiver operating characteristic and report the area under the curve, which is shown in Fig. \ref{roc}. The left one is the class-specific results of SFADA, and the right one is the overall results compared with other methods. It is obvious that (1) the automatic DR grading model trained by SFADA can perform DR diagnosis at a high threshold and (2) is better than those trained by other DA methods.

\begin{table}[htbp]
\caption{Quantitative results of LPDA and ALRM compared with other methods. The best performance is bold fonts and the second performance is underlined. All experiments are conducted with an active budget $\mathcal{B}=5\%$.}
\begin{center}
\footnotesize
\setlength{\tabcolsep}{8pt}
\begin{tabular}{@{}cccccc@{}}
\toprule
AL Method & DA Method & ACC & F1\_Score & AUC & QWK            \\ \midrule
Random & \multirow{5}{*}{LPDA} & 79.33 & 76.85 & 93.80 & 86.94          \\
AADA  & & 80.11  & 78.78  & 93.63 & {\ul 89.70}    \\
BADGE & & 78.86  & 77.49  & 92.99 & 87.36 \\
LAS  & & {\ul 81.16} & {\ul 79.75} & {\ul 94.72} & 88.75 \\
ALRM & & \textbf{85.36} & \textbf{83.71} & \textbf{95.70} & \textbf{92.38} \\ \midrule
\multirow{5}{*}{ALRM} & ft w/ CE loss & 76.20 & 73.77 & 93.22 & {\ul 85.89}    \\
& DANN & {\ul 78.78} & 75.72 & {\ul 94.66} & 84.40 \\
& MME & 78.55 & {\ul 75.93}   & 93.97& 84.97\\
& RAA & 64.87 & 62.88 & 89.75 & 76.61 \\
& LPDA & \textbf{85.36} & \textbf{83.71} & \textbf{95.70} & \textbf{92.38} \\ \bottomrule
\end{tabular}
\label{sp}
\end{center}
\end{table}
  
\subsection{Specific discussion}
Apart from the source feature generation stage, our SFADA performs active sample selection in the target domain and adaptation between source and target domains alternately. We wonder whether the proposed ALRM and LPDA are advanced in the DA tasks of DR grading. Therefore, we select some classical active sample selection methods such as Random, AADA \cite{su2020active}, BADGE \cite{ash2019deep}, LAS \cite{sun2022local} and some domain adaptation methods such as ft w/ CE, DANN \cite{ganin2016domain}, RAA \cite{sun2022local} from the related field for side-by-side comparison. The random and ft w/ CE mean randomly selecting some UWF fundus images as active samples and fine-tune the model only using active samples, respectively. LAS and RAA are the methods in our baseline. All the experimental settings are the same as quantitative evaluation. The results are presented in Table \ref{sp}. The highest score is shown in bold font, while the second-highest score is underlined. Clearly, thanks to the consideration of outliers and the full utilization of active samples, the proposed ALRM and LPDA are more suitable for domain adaptation in DR grading than other methods. In particular, the proposed methods achieve the best performance on all the overall evaluation matrices.

We conduct further study on the proposed ALRM, and the results refer to Fig. \ref{fig:budget} - Fig. \ref{fig:AL_k}. The DR grading performance of different active budgets $\mathcal{B}$ is reported in Fig. \ref{fig:budget}, which shows that (1) our ALRM have a rapid DR grading performance improvement with increasing $\mathcal{B}$ when $\mathcal{B}$ is small. This is because too few active samples cannot effectively represent the entire dataset. (2) Starting from $\mathcal{B}$=3\%, our ALRM achieves the best DR grading performance in the remaining experiments. Fig. \ref{fig:budget_fixed} is the performance of five rounds of active selection with $\mathcal{B}$=5\% and indicates that our ALRM obtains the best DR grading performance in every training epoch. We also discuss the effect of different local setting $K$ of samples on the final DR grading performance in Fig. \ref{fig:AL_k}. It demonstrates that (1) a larger $K$ in a certain range can make the proposed ALRM actively select more valuable UWF fundus images for labeling, which can effectively improve DR grading performance. However, (2) too large $K$ will make the proposed ALRM ignore some high-value UWF fundus images, which is harmful to final DR grading performance. This may be caused by the definition of local representation. In other words, the essence of local representation is the average of neighbors, and too large $K$ will cause the local representation of each sample to converge, thus leading to unsuccessful matching.

\begin{figure*}[htbp]
  \centerline{\includegraphics[scale=0.53]{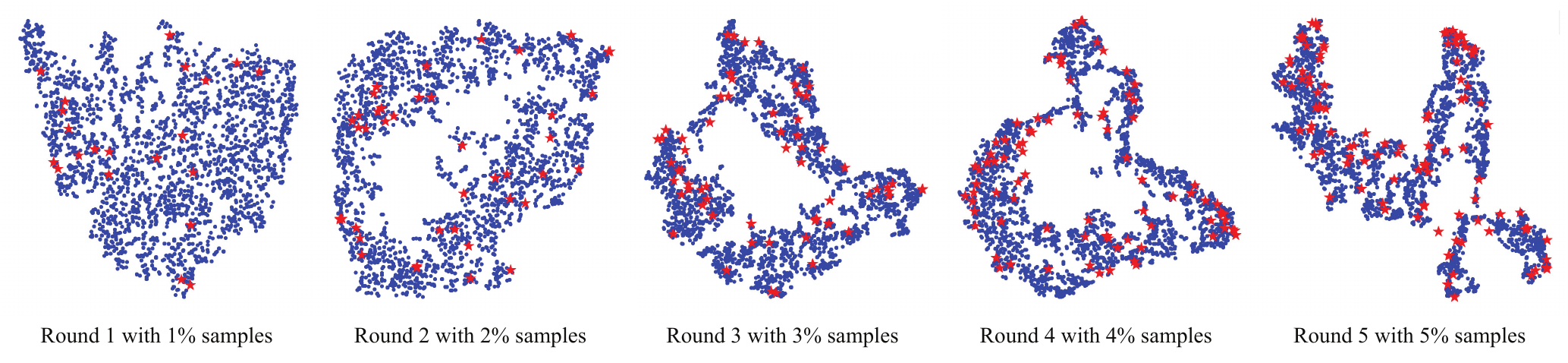}}
  \caption{Visualization of the relationship between the selected UWF fundus images (red) and all UWF fundus images (blue) in five round with 5\% budget. Local representations of each sample are reduced the dimensionality by t-SNE.}
  \label{vis_al}
  \label{vis_ll}
\end{figure*}
    
\begin{figure}[htbp]
\centerline{\includegraphics[scale=0.42]{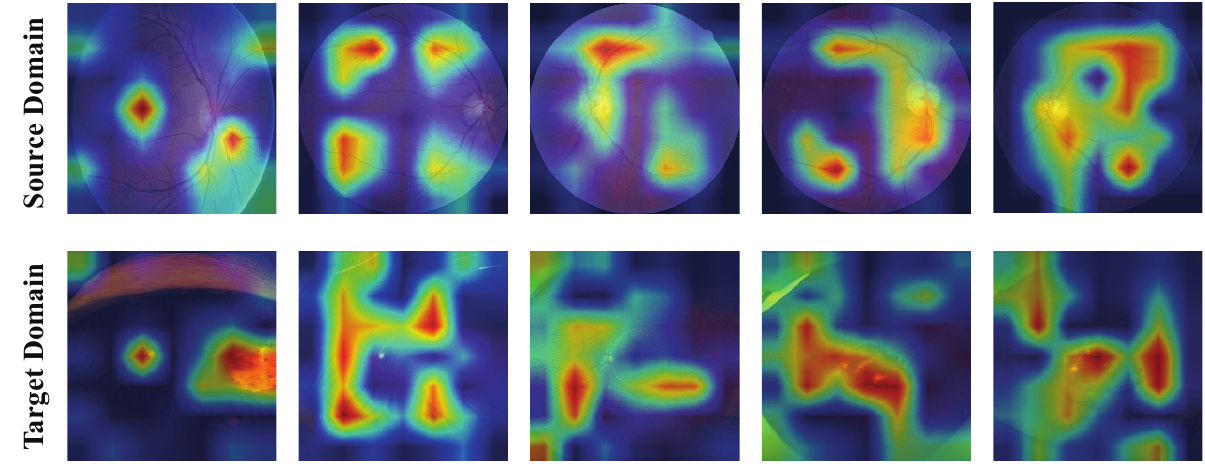}}
\caption{Visualization of the regions attended by the source pre-trained model in the source domain and the regions attended by the adapted model in the target domain. The technique of visualization is Ablation-CAM.}
\label{vis}
\end{figure}

To further investigate the reason why the proposed S-PMiS can improve rather than harm the DR grading performance, we modify the timing of pseudo-label training and conduct another four experiments to verify the accuracy of pseudo-labeling, which is shown in Fig. \ref{fig:S-RMiS}. Since our S-PMiS relies on the performance of model G, the pseudo-label accuracy generated by S-PMiS in the early stage increases rapidly with the improvement of model $G$ performance. When the performance of model $G$ reaches a certain level, the pseudo-label accuracy of S-PMiS consistently outperforms others.
  
\subsection{Visualization}
It is a fact that there are more details in the UWF fundus images that can be employed for DR grading than that in the colors fundus images \cite{ju2020bridge}. DA may focus on common knowledge to grade DR between colors fundus images and UWF fundus images, and ignore some UWF-specific information. Fortunately, active sampling with supervised training can alleviate this issue to some extent. As illustrated in Fig. \ref{vis}, We visualize the feature map of the last LayerNorm in Transformer with the help of the Ablation-CAM \cite{ramaswamy2020ablation}. (1) The interested regions focused by the source pre-trained model in the source domain color fundus images heatmap are similar to those focused by the adapted model in the target domain UWF fundus images heatmap, indicating that the meaningful knowledge is adapted to the DR grading of UWF fundus images. (2) Meanwhile, more interested regions in the heatmap of UWF fundus images than traditional color fundus images show that our SFADA captures more UWF-specific details for superior performance DR grading.

To visually demonstrate the relationship between all target domain samples and actively selected samples, we also visualize the active selection results of each round by t-SNE \cite{van2008visualizing} technology, which is shown in Fig. \ref{vis_al}. Due to the poor performance of the source pre-trained model in the UWF fundus images, the selection results of the first round are far from our expectations. With the progress of DA and the increase in labeling, the subset composed of active samples can be well matched with the whole target domain in the space of local representation.
  
\subsection{Ablation study}
The LPDA is the key to successfully adapt the source pre-training model to UWF fundus images. To investigate the impact of each component in LPDA on the final DR grading performance, six ablation experiments are designed, and the quantitative evaluation results are shown in Table \ref{ablation1}. It is clear that each of the components proposed in the LPDA contributes to improve the performance of DR grading. Specifically, the consistency regularity of unannotated UWF fundus images has the greatest improvement on the DR grading performance. Although the proposed S-PMiS improves DR grading performance less significantly than other proposed components in LPDA, it solves the problem that pseudo-labeling is harmful to the domain adaptation of DR grading in UWF fundus images, which can lead to a better performance DR grading model (for more details, refer to the next paragraph).

\begin{table}[htbp]
  \caption{Ablation study on our proposed LPDA. The best performance is \textbf{highlighted} and the second performance is \underline{underlined}. All experiments are conducted with an active budget of 5\%}
  \centering
  \footnotesize
  \setlength{\tabcolsep}{9pt}
  \resizebox{\linewidth}{!}{
  \begin{tabular}{@{}ccccc|cc@{}}
  \toprule
  $\mathcal{L}_{ce}^{tl}$ & $\mathcal{L}_{Alg}^{Inter}$ & $\mathcal{L}_{Reg}^{Inter}$ & $\mathcal{L}_{Reg}^{Intra}$ & S-PMiS & ACC & Kappa \\ \midrule
  \Checkmark & & & & &76.79 &68.74 \\
  \Checkmark &\Checkmark & & & &80.29 & 73.73 \\
  \Checkmark &\Checkmark &\Checkmark & & &83.98 &78.47     \\
  \Checkmark &\Checkmark & &\Checkmark &  &83.38 &77.80     \\
  \Checkmark &\Checkmark &\Checkmark &\Checkmark & & \underline{84.81}    &\underline{79.91}     \\
  \Checkmark &\Checkmark &\Checkmark &\Checkmark & \Checkmark &\textbf{85.36} &\textbf{80.54} \\ \bottomrule
  \end{tabular}}
  \label{ablation1}
\end{table}
  
To further explore the necessity of various designs in S-PMiS, the results of six additional ablation experiments are presented in Table \ref{ablation2}. The results indicate that (1) both pseudo-labeling and mixup compromise the DR grading performance. (2) Based on the training of mixup, the proposed Add PL, Mis PL, and Rev PL can gradually improve the DR grading performance. There is an insight to explain this issue. Although the mixup is harmful to DR grading performance, it can make the model more robust, which facilitates the identification of generated mixup samples and gives a higher accuracy pseudo label. The extra information introduced by the pseudo label can positively feed back into the training of the model, thus compensating for the performance degradation caused by robustness.
  
% \usepackage{booktabs}
\begin{table}[htbp]
  \caption{Ablation study on our proposed S-PMiS. The best performance is \textbf{highlighted} and the second performance is \underline{underlined}. All experiments are conducted with an active budget of 5\%}
  \centering
  \footnotesize
  \setlength{\tabcolsep}{9pt}
  \resizebox{\linewidth}{!}{
  \begin{tabular}{@{}cccc|cc@{}} 
  \toprule
  mixup & Add PL & MiS PL & Rev PL & ACC & Kappa \\ \midrule
  & & & &84.81 & 79.91      \\
  \Checkmark &  &  &   &84.39 &79.24\\
  \Checkmark &\Checkmark &  &  &84.43 &79.26 \\
  \Checkmark &\Checkmark &\Checkmark &  &\underline{85.08} &\underline{80.21} \\
  \Checkmark &\Checkmark &\Checkmark &\Checkmark & \textbf{85.36} &\textbf{80.54} \\
   &\Checkmark &\Checkmark &\Checkmark  & 83.79 & 78.46  \\ \bottomrule
  \end{tabular}}
  \label{ablation2}
\end{table}

\subsection{Potential limitation}
Assisting ophthalmologists in DR Diagnosis, the DR grading model trained by our SFADA may face some potential limitations, as follows: (1) Different from the unsupervised method, our SFADA requires professional ophthalmologists to provide labeling information of selected samples during the training. Although the amount of labeling is very small, it still needs the intervention of professional ophthalmologists to work well. (2) The DR grading model trained by our SFADA can not provide ophthalmologists with more detailed information about fundus lesions, such as category and location. Further analysis of UWF fundus images by professional ophthalmologists is still needed to make final conclusions. 

\section{Conclusion} \label{Conclusion}
This paper proposes a novel approach to achieve superior DR grading performance on unannotated UWF fundus images while considering data privacy and computational efficiency. Specifically, it mainly includes three parts: source feature generation, active local representation matching, and lesion-based prototype domain adaptation. Extensive experimental results show that our proposed method achieves the state-of-the-art DR grading performance and the necessity of each component. Furthermore, given the potential of the proposed framework to establish a powerful DR computer-aided diagnostic system with sustainable performance improvement, we are committed to conducting further research and exploration based on this foundation in the future, addressing more practical clinical ophthalmic issues.
 
\bibliographystyle{IEEEtran}
\bibliography{refs}{}

\end{document}